%% file: root.tex
\documentclass[letterpaper, 10 pt, conference]{ieeeconf}
\IEEEoverridecommandlockouts
\input{preamble.tex}

\makeatletter
\def\@IEEEfigurecaptionsepspace{\vskip 0pt plus 0pt minus 0pt\relax}
\makeatother

\NewDocumentCommand{\frm}{ m }{%
  \mathtt{#1}}

\NewDocumentCommand{\pos}{ O{} m m }{%
  \prescript{#1}{}{\bm{p}}_{\,#2 #3}}

\NewDocumentCommand{\rot}{ m m }{%
  {}_\mathtt{#1}\mathbf{R}_\mathtt{#2}}

\NewDocumentCommand{\rotc}{ m m }{%
  {}_{#1}\mathbf{R}_{#2}}

\NewDocumentCommand{\tf}{ m m }{%
  {}_\mathtt{#1}\mathbf{T}_\mathtt{#2}}

  \NewDocumentCommand{\tfs}{ m m m }{%
  {}_\mathtt{#1}\mathbf{T}_{\mathtt{#2}_{#3}}}

\NewDocumentCommand{\vp}{ m m  }{%
  {}_\mathtt{#2}\mathbf{#1}}

\NewDocumentCommand{\vpp}{ m m m }{%
  {}_\mathtt{#2}\mathbf{#1}_{#3}}

\NewDocumentCommand{\spp}{ m m m }{%
  {}_\mathtt{#2}{#1}_\mathtt{#3}}

\NewDocumentCommand{\spps}{ m m m m }{%
  {}_\mathtt{#2}{#1}_{\mathtt{#3}_{#4}}}

\NewDocumentCommand{\bmu}{}{%
 \bm{\mu}}

\newcommand{\so}{\textit{SO}}
\newcommand{\se}{\textit{SE}}

\title{\LARGE \bf
IMU-Preintegrated Radar Factors for Asynchronous \acl{rli} SLAM
}

\author{Johan Hatleskog, Morten Nissov, and Kostas Alexis 
\thanks{This material was supported by the Research Council of Norway under Grant 338694, Grant 310255, and Grant 321435.}
\thanks{The authors are with the Norwegian University of Science and Technology (NTNU), O. S. Bragstads Plass 2D, 7034, Trondheim, Norway {\tt\small johan.hatleskog@gmail.com}}
\thanks{Johan Hatleskog is also with Cognite AS, Lysaker, Norway}
}

\begin{document}

\maketitle
\thispagestyle{empty}
\pagestyle{empty}

\begin{abstract}
Fixed-lag \acl{rli} smoothers conventionally create one factor graph node per measurement to compensate for the lack of time synchronization between radar and LiDAR. For a radar-LiDAR sensor pair with equal rates, this strategy results in a state creation rate of twice the individual sensor frequencies. This doubling of the number of states per second yields high optimization costs, inhibiting real-time performance on resource-constrained hardware. We introduce \emph{IMU-preintegrated radar factors} that use high-rate inertial data to propagate the most recent LiDAR state to the radar measurement timestamp. This strategy maintains the node creation rate at the LiDAR measurement frequency. Assuming equal sensor rates, this lowers the number of nodes by \qty{50}{\percent} and consequently the computational costs. Experiments on a single board computer (which has 4 cores each of \qty{2.2}{\giga\hertz} A73 and \qty{2}{\giga\hertz} A53 with \qty{8}{\giga\byte} RAM) show that our method preserves the absolute pose error of a conventional baseline while simultaneously lowering the aggregated factor graph optimization time by up to \qty{56}{\percent}.
\end{abstract}


\section{Introduction}\label{sec:intro}
Accurate and reliable state estimation is a prerequisite for autonomous navigation in GPS-denied and perceptually challenging environments. Recently, Frequency-Modulated Continuous Wave (FMCW) radar has gained increasing attention as a complementary sensor to LiDAR. FMCW radar systems offer direct Doppler velocity measurements and operate robustly in degraded visual conditions such as dust, fog, rain, or darkness as well as geometrically uninformative environments such as tunnels or open spaces \cite{nissov2024degradationIcra}. When fused with LiDAR and inertial measurements, radar measurements can improve resilience in scenarios where LiDAR-only approaches degrade.

In factor graph-based smoothing for multi-modal sensor fusion, the conventional method to address measurement asynchronicity is to create a distinct state node in the graph for each measurement from each asynchronous sensor. For a typical \acl{rli} setup with a \qty{10}{\hertz} radar and a \qty{10}{\hertz} LiDAR, this results in 20 state nodes being added to the graph every second. The theoretic computational complexity of factor graph optimization scales quadratically with the number of states in the optimization window \cite{dellaert2017factor}. Consequently, this doubling of the state creation rate imposes a significant computational burden, which can inhibit real-time performance, especially on the resource-constrained, embedded systems commonly found on mobile robots.

To address this limitation, we propose a principled approach that avoids creating state nodes for radar measurements. Instead, we introduce \emph{IMU-preintegrated radar factors}. Our method creates state nodes only upon the arrival of a LiDAR measurement. When a radar measurement is received, we use high-rate IMU data to propagate the state from the preceding LiDAR node to the timestamp of the radar measurement. This propagation yields a predicted state that is used to evaluate the radar factor residual, creating a constraint on the previous LiDAR node without adding a new state to the graph.

\begin{figure}[t]
    \centering
    \includegraphics[width=\linewidth]{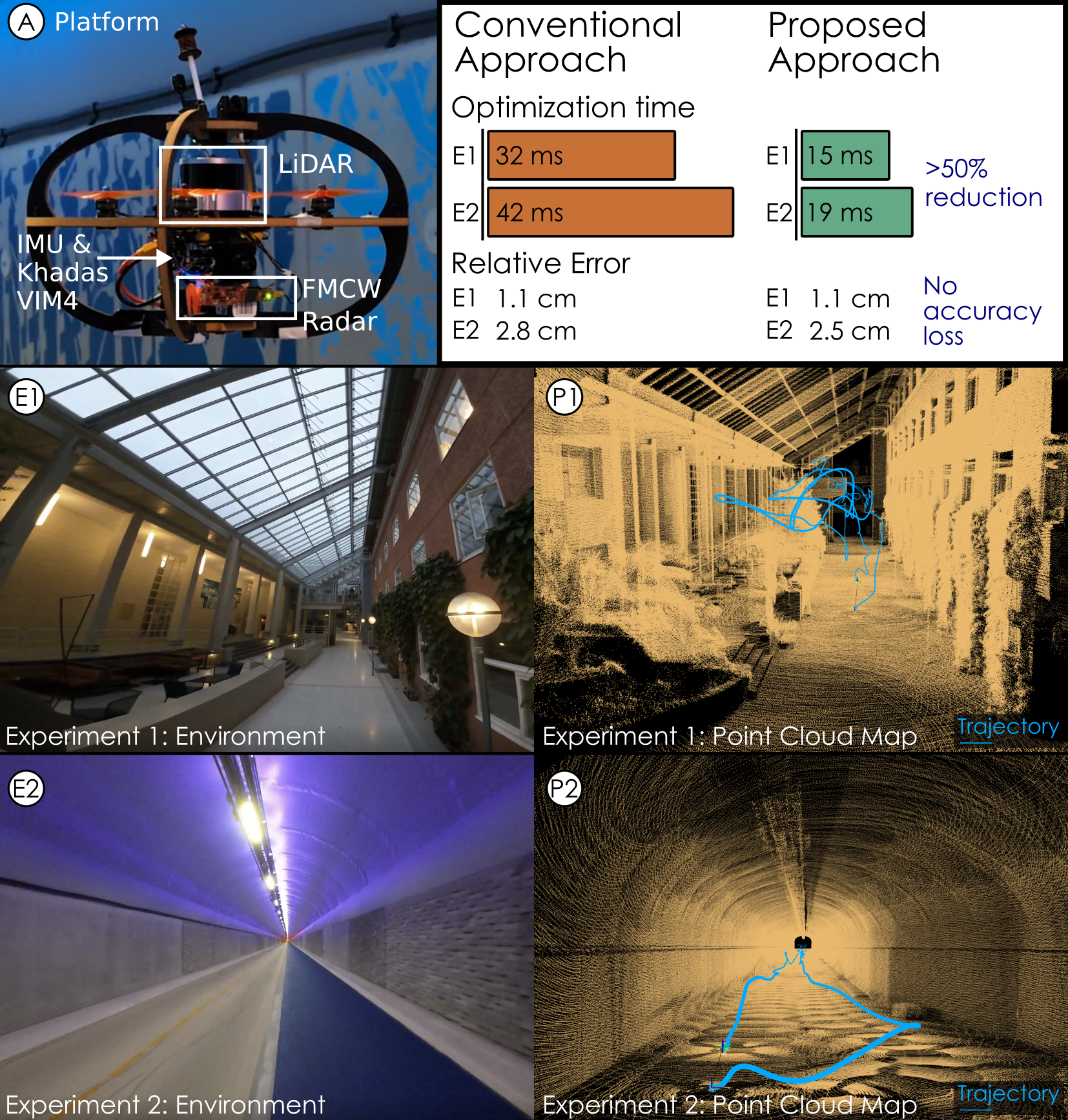}
    \caption{An overview of the results and environments considered in this manuscript, including (A) the aerial platform flying in the Fyllingsdalen tunnel environment, (E1) an image and (P1) point cloud map from the NTNU environment, and (E2) an image and (P2) point cloud map from the Fyllingsdalen tunnel environment.}
    \label{fig:fpf}
    \vspace{-1em}
\end{figure}

With \qty{10}{\hertz} radar-LiDAR sensor rates, this strategy reduces the number of nodes in the fixed-lag smoother by \qty{50}{\percent} compared to the standard approach. Our main contributions are:
\begin{enumerate}
    \item A new IMU-preintegrated radar factor that removes the need for dedicated radar state nodes.
    \item A computationally efficient factor graph smoother formulation, utilizing the IMU-preintegrated radar factor, suitable for real-time \acl{rli} fusion on resource-constrained embedded hardware.
    \item Experimental evaluations on real-world datasets on constrained hardware, demonstrating that our approach substantially reduces computational load with a negligible effect on localization accuracy compared to a conventional baseline.
\end{enumerate}
The aerial platform and the environments in which the experiment took place are depicted in Fig. \ref{fig:fpf}~(A) and Fig. \ref{fig:fpf}~(E1-E2), respectively. A supplementary video of the experimental results accompanies this paper: \mbox{\url{https://youtu.be/95jeWXBMN7c}}.

The remainder of this work is organized as follows: Section \ref{sec:related} presents related work and the proposed approach is described in Section \ref{sec:approach}. The evaluation study is presented in Section \ref{sec:evaluation} and we conclude in Section \ref{sec:conclusions}.

\section{Related Work}\label{sec:related}

This work addresses computationally efficient, asynchronous \acl{rli} odometry. Our method relates to LiDAR-inertial odometry, \acl{rli} fusion and methods for handling asynchronous sensor measurements in factor graph-based sensor fusion.

\subsection{LiDAR-Inertial Odometry}

LiDAR-inertial odometry (LiO) is widely used for robust localization and mapping. We briefly review a subset of the related work in LiDAR-inertial odometry, and refer to \cite{lee2024lidar} for a comprehensive overview. A large body of work builds on LOAM \cite{zhang2014loam}, a seminal loosely coupled LiDAR inertial framework. To improve accuracy, subsequent LiO frameworks tightly couple LiDAR and IMU measurements using extended Kalman filters \cite{xu2022fast} or factor graph optimization \cite{shan2020lio}. Frameworks such as \cite{khedekar2022mimosa,zhao2021super} enable factor graph-based fusion of LiDAR, IMU and complementary modalities. The factor graph-based frameworks create state nodes in the graph for each non-IMU measurement, and use IMU preintegration \cite{forster2016manifold} to summarize intra-node IMU-measurements into a single relative motion factor between consecutive state nodes. Our work builds on the factor graph-based approaches but focuses on adding an additional asynchronous sensor modality at a limited computational cost by extending the use of IMU preintegration.

\subsection{\acl{rli} Fusion}

Radar is a compelling complementary modality to LiDAR in perceptually degraded environments \cite{nissov2024degradationIcra}. In geometrically uninformative environments, radar can provide information in directions of the state space that are unobservable from LiDAR-IMU alone. Radar also offers resilience to obscurants such as dust and fog \cite{fritsche2018fusing,kramer2020radar} due to the longer wavelength. We refer to \cite{harlow2024new} for a comprehensive review of \unit{\milli\meter}Wave radar usage in robotics. Our work focuses on Frequency Modulated Continuous Wave (FMCW) mmWave radar sensing, whose per-point Doppler radial velocity measurements provide direct measurements of the sensor's ego-velocity in a static environment. Related work on radar-inertial fusion include loosely coupled \cite{doer2021x} and tightly-coupled \cite{michalczyk2022tightly} filtering, as well as factor graph based smoothing \cite{kramer2020radar}. Similarly, \acl{rli} fusion have been addressed with filtering \cite{fritsche2018fusing,noh2025garlio} and factor graph-based smoothing \cite{nissov2024degradationIcra,nissov2024roamer,nissov2024robust} approaches, where we build on the latter class of works. Compared to works such as \cite{nissov2024roamer,nissov2024robust}, where graph nodes are created at the rate of the IMU to facilitate adding non-synchronized measurements, we reduce the computational cost of the smoother optimization by only creating state nodes for LiDAR measurements and applying the proposed factor for radar measurements. Works such as \cite{nissov2024degradationIcra,noh2025garlio} address the problem by deskewing the LiDAR point cloud to the radar timestamps, thus avoiding the need for additional nodes in the graph. Our approach offers the same computational benefits without coupling the radar and LiDAR preprocessing, offering additional robustness with respect to potential sensor dropout.

\subsection{Asynchronous Graph-Based Sensor Fusion}

Dealing with time-asynchronous measurements is a pervasive challenge for factor graph-based multi-sensor fusion. The conventional approach in a discrete-time factor graph is to create a new state node for every measurement \cite{indelman2013information}. Although accurate and conceptually simple, this method has the drawback of inflating the number of variables. This significantly increases the computational cost, which is polynomial in the number of nodes \cite{dellaert2017factor}. A practical solution which does not result in the same computational cost is to interpolate poses or measurements to a common timestamp \cite{geneva2018asynchronous}, though this may not accurately capture non-linear system dynamics.

Continuous-time SLAM offers an elegant solution to asynchronous sensor fusion \cite{droeschel2018efficient,furgale2012continuous}. By representing the trajectory as a continuous function, often using B-splines, measurements can be incorporated at their precise timestamps. However, this approach introduces a critical trade-off: choosing the temporal density of the trajectory representation to balance real-time feasibility and accuracy is a non-trivial and open problem \cite{talbot2024continuous}. In contrast, our method avoids this challenge by creating states at the rate of the primary sensor, offering a practical and real-time feasible solution for resource-constrained hardware.

The approach most related to ours is that of output prediction. In a filtering context,  \cite{khosravian2015recursive} introduced output prediction for recursive attitude estimation on the $\so(3)$ Lie group of 3D rotations. In this method, a high-frequency motion model propagates the state estimate forward to the arrival time of a delayed measurement in order to create the arrival time measurement prediction, which is then fused in the observer. Our proposed IMU-preintegrated factor applies this same idea within a factor graph-based smoothing estimator to relate incoming radar measurements to a nearby LiDAR measurement anchor. We use on-manifold IMU preintegration \cite{forster2016manifold} to derive a non-linear constraint on a past state from an asynchronous radar measurement. This method avoids the computational cost of creating new states while accurately modelling the system dynamics between the state and the measurement.

\section{Proposed Approach}\label{sec:approach}

\begin{figure*}
    \centering
    \vspace{0.5mm}
    \includegraphics[width=\linewidth]{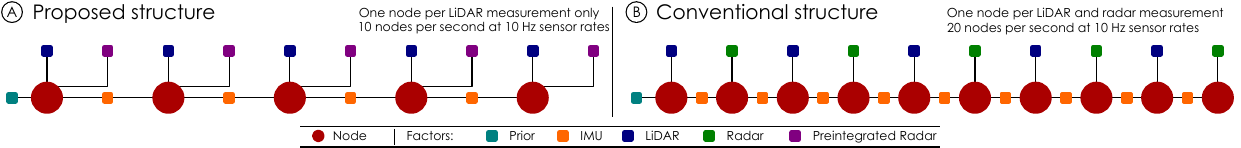}
    \caption{Factor graph fixed-lag smoother structure with (A) our proposed approach with one node per LiDAR measurement only and (B) the conventional approach with one node per LiDAR and radar measurement. At \qty{10}{\hertz} sensor rates, our approach reduces the amount of nodes needed by \qty{50}{\percent} compared to the conventional approach, resulting in computational savings and reduced latency.}
    \label{fig:method}
    \vspace{-1em}
\end{figure*}

This section presents the proposed preintegrated radar factor and the fixed-lag factor graph-based \acl{rli} smoothing framework in which it is used.

\subsection{Notation and frames}

This work uses the following notation. Scalars are in normal case (e.g. $x$), while vectors and matrices are boldfaced (e.g. $\mathbf{x}$, $\mathbf{X}$). The pose of a frame $\frm{A}$ with respect to frame $\frm{B}$ is represented by the homogeneous transformation $\tf{B}{A} \in \se(3)$, which consists of a rotation matrix $\rot{B}{A} \in \so(3)$ and a translation vector $\vpp{p}{B}{\frm{BA}} \in \mathbb{R}^3$. The $\text{Log}(\cdot)$ operator is used to map Lie group members (e.g., $\so(3)$ and $\se(3)$) to their corresponding Lie algebras. We refer to \cite{sola2018micro} for an introduction to Lie theory. The body-fixed frames for the IMU, radar, and LiDAR are denoted $\frm{I}$, $\frm{R}$, and $\frm{L}$, respectively. All poses are estimated with respect to a static world frame $\frm{W}$. The extrinsic transformations between sensor frames, $\tf{I}{L}$ and $\tf{I}{R}$, are assumed to be known from prior calibration.

\subsection{System State and Factor Graph Structure}
Our system employs a fixed-lag smoother, where the states within the sliding window encompassing the $l$ most recent measurements are estimated using a factor graph. We create state nodes in the factor graph only at the timestamps of incoming LiDAR measurements and connect them with preintegrated IMU measurements, following \cite{forster2016manifold}. The state $\mathbf{x}_i$ at time $t_i$ is defined as
\begin{equation}
    \mathbf{x}_i = (\tfs{W}{I}{i},\ \vpp{v}{W}{\frm{WI}_i},\ \mathbf{b}_i)
\end{equation}
where $\tfs{W}{I}{i}$ is the pose of the IMU in the world frame, $\vpp{v}{W}{\frm{WI}_i}$ is the linear velocity of the IMU in the world frame, and ${\mathbf{b}_{i} = [\mathbf{b}_{ai}^T, \mathbf{b}_{gi}^T]^T \in \mathbb{R}^6}$ contains the accelerometer and gyroscope biases, respectively. The unit-length gravity direction vector $\vp{{\Bar{g}}}{W}$ is also estimated, assuming the local gravity magnitude $g$ is known a-priori, such that the gravity vector is ${\vp{{{g}}}{W} = g \vp{{\Bar{g}}}{W}}$.

The factor graph smoother estimates the states $\mathbf{x}_{k-l:k} = \{\mathbf{x}_{k-l}, \dots, \mathbf{x}_k \}$ as the optimizer
\begin{equation}
\begin{split}
    \mathbf{x}_{k-l:k}^\star = \arg \min_{\mathbf{x}_{k-l:k}} \Biggl( &\sum_{i=k-l}^k \Big[ \lVert \mathbf{e}_{\mathcal{L}_i} \rVert_{\mathbf{\Sigma}_{\mathcal{L}_i}}^2 + \lVert \mathbf{e}_{\mathcal{R}_i} \rVert_{\mathbf{\Sigma}_{\mathcal{R}_i}}^2 \Big]  \\ &+\sum_{i=k-l}^{k-1} \lVert \mathbf{e}_{\mathcal{I}_i} \rVert_{\mathbf{\Sigma}_{\mathcal{I}_i}}^2 + \lVert \mathbf{e}_{\mathcal{P}} \rVert_{\mathbf{\Sigma}_\mathcal{P}}^2 \Biggl),
    \end{split}
\end{equation}
where $\mathbf{e}_{\mathcal{L}_i}$, $\mathbf{e}_{\mathcal{R}_i}$, $\mathbf{e}_{\mathcal{I}_i}$ are residuals from LiDAR, radar and IMU measurements respectively and $\mathbf{\Sigma}_{\mathcal{L}_i}$,$\mathbf{\Sigma}_{\mathcal{R}_i}$ and $\mathbf{\Sigma}_{\mathcal{I}_i}$ are their covariances. $\mathbf{e}_{\mathcal{P}}$ is the marginalization prior residual. The factor graph structure is exemplified in Fig.~\ref{fig:method}~(A). We describe the individual factors in the following sections. The Jacobians of the factors are omitted for brevity.

\subsection{Degeneracy-Aware LiDAR Factors}\label{ssec:lidar_factors}
At each state node time $t_i$, a LiDAR measurement is incorporated as a unary factor constraining the IMU pose $\tfs{W}{I}{i}$. The measurement, denoted ${}_\frm{W}\tilde{\mathbf{T}}_{\frm{L}_i} \in \se(3)$, is obtained from the degeneracy-aware scan-to-map registration framework described in \cite{hatleskog2024probabilistic}. The LiDAR factor residual is
\begin{equation}
    \mathbf{e}_{\mathcal{L}_i} = \text{Log}\left( ( {}_\frm{W}\tilde{\mathbf{T}}_{\frm{L}_i} \tf{L}{I} )^{-1} \tfs{W}{I}{i} \right).
    \label{eq:lidar_residual}
\end{equation}
The framework importantly provides a degeneracy-aware covariance estimate $\mathbf{\Sigma}_{\mathcal{L}_i} \in \mathbb{R}^{6\times6}$ associated with the measurement. The covariance estimate is inflated in directions identified as geometrically degenerate, such as the translation along a long corridor, allowing complementary modalities to inform directions that are unobservable from LiDAR measurements alone.

\subsection{IMU Factors}
\label{ssec:imu_factors}
To constrain the motion between consecutive state nodes $(\mathbf{x}_i, \mathbf{x}_{i+1})$, we create an IMU factor using the preintegration scheme of \cite{forster2016manifold}. This method summarizes the high-frequency IMU measurements between the node timestamps $t_i$ and $t_{i+1}$ into a single relative motion constraint, avoiding the cost of adding states at the IMU rate.

The integration of raw accelerometer $\tilde{\mathbf{a}}_k$ and gyroscope $\tilde{\bm{\omega}}_k$ measurements between $t_i$ and $t_{i+1}$ yields a preintegrated measurement $\Delta\tilde{\bm{\chi}}_{i,i+1} = (\Delta\tilde{\mathbf{R}}_{i,i+1}, \Delta\tilde{\mathbf{v}}_{i,i+1}, \Delta\tilde{\mathbf{p}}_{i,i+1})$, from which the change in orientation, velocity, and position between $\mathbf{x}_i$ and $\mathbf{x}_{i+1}$ can be derived.  

The preintegration scheme assumes that the bias is constant over the preintegration interval, and the initial preintegrated measurement $\Delta\tilde{\bm{\chi}}_{i,i+1}$ is computed  using the bias estimates available at the time of integration. During optimization, as the bias estimates $\mathbf{b}_{i}$ are updated, the the preintegrated measurements are corrected using a first-order Taylor expansion rather than performing costly re-integration. The corrected preintegrated measurement $\Delta\hat{\bm{\chi}}_{i,i+1} =(\Delta\hat{\mathbf{R}}_{i,i+1}, \Delta\hat{\mathbf{v}}_{i,i+1}, \Delta\hat{\mathbf{p}}_{i,i+1})$ is a function of the accelerometer and gyroscope biases.

The IMU factor residual, $\mathbf{e}_{\mathcal{I}_i}$, compares the bias-corrected preintegrated measurement $\Delta\hat{\bm{\chi}}_{i,i+1}$ to the change in orientation, position and velocity between states $\mathbf{x}_i$ and $\mathbf{x}_{i+1}$ in the factor graph. The components of $\mathbf{e}_{\mathcal{I}_i}$ are defined as
\begin{align}
    \mathbf{e}_{\mathcal{I}_i}^\mathbf{R} &= \text{Log}\left( (\Delta\hat{\mathbf{R}}_{i,i+1})^\top \rotc{\frm{I}_i}{\frm{W}} \rotc{\frm{W}}{\frm{I}_{i+1}} \right) \\
    \mathbf{e}_{\mathcal{I}_i}^\mathbf{p} &= \rotc{\frm{I}_i}{\frm{W}} \left( \vpp{p}{W}{\frm{WI}_{i+1}} \!-\! \vpp{p}{W}{\frm{WI}_i} \!-\! \vpp{v}{W}{\frm{WI}_i}\Delta t_i \!-\! \frac{1}{2}\vp{{g}}{W}\Delta t_i^2 \right) \nonumber \\
    & \qquad - \Delta\hat{\mathbf{p}}_{i,i+1} \\
    \mathbf{e}_{\mathcal{I}_i}^\mathbf{v} &= \rotc{\frm{I}_i}{\frm{W}} \left( \vpp{v}{W}{\frm{WI}_{i+1}} \!-\! \vpp{v}{W}{\frm{WI}_i} \!-\! \vp{{g}}{W}\Delta t_i \right) - \Delta\hat{\mathbf{v}}_{i,i+1} \\
    \mathbf{e}_{\mathcal{I}_i}^\mathbf{b} &= \mathbf{b}_{i+1} - \mathbf{b}_{i}
\end{align}
where $\Delta t_i \!=\! t_{i+1} \!-\! t_i$. The associated covariance $\mathbf{\Sigma}_{\mathcal{I}_i}$ is computed upon preintegration. The bias residual $\mathbf{e}_{\mathcal{I}_i}^\mathbf{b}$ models the IMU biases as a random walk process.

\subsection{FMCW Radar Velocity Factors}

The radar measurements are used for radar velocity factors that constrain the linear velocity $\vpp{v}{W}{\frm{WI}_i}$ of the IMU. This section first describes the measurement model. It then details the standard formulation used as a baseline, followed by our proposed IMU-preintegrated factor. The presentation of the measurement model and baseline factor follows \cite{nissov2024degradationIcra}.

\subsubsection{Radar Velocity Measurement Model}
\label{sssec:radar_model}
FMCW radar provides a sparse point cloud where each point is augmented with its measured radial velocity $\tilde{v}_r$ relative to the sensor. For a stationary point in the environment, the true radial velocity $v_r$ is the projection of the radar's ego-velocity, $\vpp{v}{R}{\frm{WR}}$, onto the point's bearing vector, $\vp{\bmu}{R}$, such that
\begin{equation}
    v_r = -\vp{\bmu}{R}^\top \vpp{v}{R}{\frm{WR}}.
\end{equation}
This motivates the definition of a single-point radar velocity residual, $\mathbf{e}_\mathcal{R}$, comparing the predicted and measured radial velocities:
\begin{equation}
    \mathbf{e}_\mathcal{R} = -\vp{\bmu}{R}^\top \vpp{v}{R}{\frm{WR}} - \tilde{v}_r.
    \label{eq:residual-radar-generic}
\end{equation}
To evaluate this residual within the factor graph, the radar's ego-velocity $\vpp{v}{R}{\frm{WR}}$ is expressed as a function of the state variables and IMU gyroscope measurement $\spp{\tilde{\bm{\omega}}}{I}{WI}$ at the time of the radar measurement:
\begin{equation}
    \vpp{v}{R}{\frm{WR}} = \rot{R}{I} \left( \rot{W}{I}^\top \vpp{v}{W}{\frm{WI}} + (\spp{\tilde{\bm{\omega}}}{I}{WI} - \mathbf{b}_g) \times \vpp{p}{I}{\frm{IR}} \right).
    \label{eq:radar-ego-vel-from-state}
\end{equation}

\subsubsection{Standard Factor}
\label{sssec:standard_radar_factor}
The standard method for incorporating an asynchronous radar measurement in a factor graph, is to create a dedicated state node $\mathbf{x}_i$ at the timestamp $t_i$ of the radar point cloud and evaluate the residual \eqref{eq:residual-radar-generic} using $\mathbf{x}_i$ with \eqref{eq:radar-ego-vel-from-state}. For a radar point cloud received at time $t_i$ with $N_{\mathcal{R}_i}$ points, the full aggregated residual is
\begin{align}
    \mathbf{e}_{\mathcal{R}_i}^{\text{bl}} = \begin{bmatrix}
        \mathbf{e}_{\mathcal{R}}(\vpp{v}{R}{\frm{WR}}(\mathbf{x}_i; \spp{\tilde{\bm{\omega}}}{I}{WI_{i}}); {}_{\mathtt{R}}\bm{\mu}_{i1}, \tilde{v}_{ri1}) \\ \vdots \\ \mathbf{e}_{\mathcal{R}}(\vpp{v}{R}{\frm{WR}}(\mathbf{x}_i; \spp{\tilde{\bm{\omega}}}{I}{WI_{i}}); {}_{\mathtt{R}}\bm{\mu}_{iN_{\mathcal{R}_i}}, \tilde{v}_{riN_{\mathcal{R}_i}})
    \end{bmatrix}, \label{eq:radar-residual-baseline}
\end{align}
with an associated isotropic covariance matrix $\mathbf{\Sigma}_{\mathcal{R}}^{\text{bl}} = \mathbf{I} \sigma_r^2$.

While straightforward, this approach inflates the node count in the factor graph. For instance, using a \qty{10}{\hertz} LiDAR and a \qty{10}{\hertz} radar results in 20 nodes per second, a doubling from the 10 nodes per second used in standard LiDAR-inertial smoothing. The inflation of nodes per second significantly increases the number of variables in the optimization window, adversely affecting computational performance. 

The factor graph structure exemplified in Fig. \ref{fig:method}~(B) and the residual \eqref{eq:radar-residual-baseline} is used as the baseline we compare against in the experimental evaluations.

\subsubsection{Proposed IMU-Preintegrated Factor}
\label{sssec:preintegrated_radar_factor}
Our main contribution is a method that avoids creating dedicated radar state nodes. Consider a radar point cloud timestamped at time $t_{ir}$, such that it falls after the LiDAR-based state node $i$ at time $t_i$ (i.e., $t_i \le t_{ir}$). Our proposed factor constrains the state $\mathbf{x}_i$ rather than creating a dedicated state node at time $t_{ir}$. This is achieved by using IMU measurements to propagate $\mathbf{x}_i$ to the radar timestamp $t_{ir}$. 

Using the same preintegration scheme described in Section~\ref{ssec:imu_factors}, we summarize the IMU measurements between $t_i$ and $t_{ir}$ into the preintegrated measurement $(\Delta\tilde{\mathbf{R}}_{i,ir},\Delta\tilde{\mathbf{v}}_{i,ir})$. Analogous to the IMU factor, these preintegrated measurements are corrected at optimization time using the current estimate of the bias $\mathbf{b}_{i}$. This yields the corrected preintegrated rotation $\Delta\hat{\mathbf{R}}_{i,ir}$ and velocity $\Delta\hat{\mathbf{v}}_{i,ir}$. We use the corrected preintegrated quantities to predict the IMU orientation and velocity at the radar timestamp $t_{ir}$:
\begin{align}
    \spps{\hat{\mathbf{R}}}{W}{I}{ir} &= \rotc{\frm{W}}{\frm{I}_i}  \Delta\hat{\mathbf{R}}_{i,ir} \label{eq:prop_rot},\\
    \spps{\hat{\mathbf{v}}}{W}{WI}{ir} &= \vpp{v}{W}{\frm{WI}_i} + \vp{{g}}{W} (t_{ir} \!-\! t_i) + \rotc{\frm{W}}{\frm{I}_i} \Delta\hat{\mathbf{v}}_{i,ir}. \label{eq:prop_vel}
\end{align}
Using the propagated variables \eqref{eq:prop_rot}-\eqref{eq:prop_vel} with \eqref{eq:residual-radar-generic}-\eqref{eq:radar-ego-vel-from-state} and the bearing vector and radial speed $({}_{\mathtt{R}}\bm{\mu}_{ij}, \tilde{v}_{rij})$ of the $j$'th point in the radar point cloud yields the per-point IMU-preintegrated radar residual
\begin{align}
    \mathbf{e}_{\mathcal{R}_{ij}} = -{}_{\mathtt{R}}\bm{\mu}^\top_{ij} \Bigg[ &\rot{R}{I} \bigg(  (\spps{\hat{\mathbf{R}}}{W}{I}{ir})^\top \spps{\hat{\mathbf{v}}}{W}{WI}{ir} \nonumber \\ 
    & \left. + (\spps{\tilde{\bm{\omega}}}{I}{WI}{{ir}} \!-\! \mathbf{b}_{gi}) \times \vpp{p}{I}{\frm{IR}} \bigg) \right] - \tilde{v}_{rij}, \label{eq:residual-preint-radar}
\end{align}
where $\spps{\tilde{\bm{\omega}}}{I}{WI}{ir}$ is the gyroscope measurement at time $t_{ir}$. Since the bias is assumed to be constant between subsequent nodes $(\mathbf{x}_i,\mathbf{x}_{i+1})$, the gyroscope bias $\mathbf{b}_{gi}$ is used without modifications. The final preintegrated radar residual $\mathbf{e}_{\mathcal{R}_i}$ stacks all the $N_{\mathcal{R}_i}$ point-wise residuals from the $i$'th sparse radar point cloud:
\begin{align}
    \mathbf{e}_{\mathcal{R}_i} = \begin{bmatrix}
        \mathbf{e}_{\mathcal{R}_{i1}} \cdots \
        \mathbf{e}_{\mathcal{R}_{iN_{\mathcal{R}_i}}}
    \end{bmatrix}^T.
\end{align}
The associated covariance $\mathbf{\Sigma}_{\mathcal{R}_i}$ is the sum of the user-defined radar measurement variance $\sigma_r^2$ and first-order noise propagation of the estimated covariance on the preintegrated quantities $(\Delta\tilde{\mathbf{R}}_{i,ir},\Delta\tilde{\mathbf{v}}_{i,ir})$. Analogous to the IMU factor, the covariance of the preintegrated quantities is computed upon preintegration.

The factor depends only the past state $\mathbf{x}_i$ and the gravity vector, thus creating a radar velocity constraint without the need to create a node at the radar timestamp.

 Since the IMU-preintegated radar factor connected to state $\mathbf{x}_i$ and an IMU factor connecting states $(\mathbf{x}_i,\mathbf{x}_{i+1})$ use a subset of the same IMU-measurements, there will be some correlation among them. The correlation is assumed to have a minor effect, and thus is neglected in the proposed formulation in favour of simplicity and improved computational efficiency.

\subsection{Prior Factor}
\label{sssec:prior_factor}

The information from the nodes exiting the smoother window is preserved through a prior factor that constrains the oldest state in the window, $\mathbf{x}_{k_0}$ and the gravity direction vector. The prior factor residual is
\begin{equation}
    \mathbf{e}_{\mathcal{P}} = 
    \begin{bmatrix}
        \text{Log}(\tfs{W}{I}{p}^{-1} \tfs{W}{I}{k_0}) \\
        \vpp{v}{W}{\frm{WI}_{p}} - \vpp{v}{W}{\frm{WI}_{k_0}} \\
        \mathbf{b}_{p} - \mathbf{b}_{k_0} \\
        \vp{\Bar{g}}{W}_p \ominus \vp{\Bar{g}}{W}
    \end{bmatrix},
\end{equation}
where $\ominus$ computes the error on the unit sphere $S^2$, yielding a 2D vector in the tangent plane at the linearization point. The linearization point $(\tf{W}{I_{p}}, \vpp{v}{W}{\frm{WI}_{p}}, \mathbf{b}_{p}, \vp{\Bar{g}}{W}_p)$ and the covariance $\mathbf{\Sigma}_{\mathcal{P}}$ associated with the prior is obtained from the marginalization procedure.

\section{Evaluation Studies}\label{sec:evaluation}
The evaluation is designed to investigate if the proposed approach yields localization accuracy similar to that of a representative baseline at a lower computational cost. To that end, we simultaneously compare the wall time consumption and positional errors of our proposed approach to a baseline approach in which factor graph nodes are created for both LiDAR and radar measurements. 

The experimental data is captured by a custom quadrotor aerial platform, shown in Fig.~\ref{fig:fpf}~(A), carrying VectorNav VN-100 IMU (\qty{200}{\hertz}), Ouster OS0-128 LiDAR (\qty{10}{\hertz}), and Texas Instruments IWR6843AOP FMCW radar (\qty{10}{\hertz}) sensors. To collect the experimental data, this aerial platform is flown, manually, through a university corridor at NTNU in Trondheim, Norway (referred to as experiment 1) in addition to the Fyllingsdalen bicycle tunnel in Bergen, Norway (referred to as experiment 2). In both environments, the platform is tracked by a Leica Geosystems MS60 MultiStation, the data from which will be used for position ground truth to create numerical results for evaluating the performance of the proposed method.

\subsection{Evaluation Metrics}
To assess the performance of our approach versus the baseline, we compare 1) the absolute (ATE) and relative (RTE) trajectory errors of the trajectories with respect to the ground truth using EVO \cite{grupp2017evo}; and 2) the wall time used for factor graph optimization. The RTE is evaluated per 1 meter traversed. We report three wall clock metrics: 1) the total wall time for a full experiment; 2) the average wall time per \qty{100}{\milli\second} (i.e., total time divided by the number of \qty{100}{\milli\second} ticks); and 3) the wall time per factor graph iteration.

\subsection{Implementation}
The fixed-lag smoother described in section \ref{sec:approach} is implemented in C++ using the open-source SymForce library \cite{martiros2022symforce} for factor graph optimization and symbolic code-generation of the aforementioned residuals and their Jacobians. The implementation is single-threaded. The Huber robust cost function \cite{huber1964robust} is applied to the per-point radar velocity residuals to mitigate the effect of outliers.

For a fair comparison, the baseline approach, with one node per LiDAR and radar, is implemented within the same framework as our proposed approach. For the baseline approach, the graph is optimized both upon processing LiDAR and radar measurements. This is necessary to keep the number of nodes in the graph fixed, as the oldest node is marginalized out upon adding a new node, and additionally benefits the accuracy of the baseline by updating the estimate with the latest available measurements. In our proposed approach, the graph is optimized only upon processing LiDAR measurements. The computational load scales with the number of factor graph iterations per optimization. Each optimization terminates when the relative reduction in error falls below a threshold. To assess the performance implications of using radar in addition to LiDAR-inertial measurements, we also include reference metrics from the LiO method resulting from using the same implementation as our proposed approach, but without any radar factors. In all cases, the framework outputs state estimates at the IMU rate by using the history of recent IMU messages to propagate the latest factor graph optimized estimate up to the time of the most recent IMU message.

The LiDAR measurements for the factor in section \ref{ssec:lidar_factors} are obtained by using the scan-to-map registration framework described in \cite{hatleskog2024probabilistic}. Each LiDAR scan is deskewed, using a constant acceleration and angular rate model between successive IMU measurements, followed by LOAM style feature extraction \cite{zhang2014loam} and additional voxel filtering. The extracted and downsampled features are then used for scan-to-map registration to a monolithic point cloud map. The pose resulting from scan-to-map optimization is passed to the factor graph smoother, and the deskewed and downsampled scan is integrated into the map by using the refined pose estimate from factor graph optimization. The initial guess for scan-to-map registration and intra-sweep motion estimates for deskewing are computed using the IMU-rate estimates from the factor graph. 

Furthermore, the full resolution deskewed LiDAR point cloud is output to facilitate downstream mapping and planning with greater detail than necessary for the registration in the estimator. In this manuscript, we use the full-resolution deskewed point clouds with their associated pose estimates from the factor graph optimization to build a dense map offline for visualization purposes.

To demonstrate the utility of the proposed approach on constrained hardware relevant for use on drones, we run the experiments on a Khadas VIM4 with \qty{8}{\giga\byte} RAM and a combined \qty{2.2}{\giga\hertz} quad-core Cortex-A73 CPU and a \qty{2.0}{\giga\hertz} quad-core Cortex-A53 CPU. To allow for real-time operation, the smoother window is fixed to 2.5 seconds. For our proposed approach, we create nodes at the \qty{10}{\hertz} LiDAR rate and thus the smoother window duration corresponds to 25 nodes in the graph. For the same smoother window size the baseline approach would create a graph with 50 nodes of the interleaved \qty{10}{\hertz} LiDAR and \qty{10}{\hertz} radar measurements.

\subsection{Experiment 1: NTNU Elektro}\label{sec:evaluation:elektro}

\begin{figure*}
\centering
\vspace{0.5mm}
\includegraphics[width=1\textwidth]{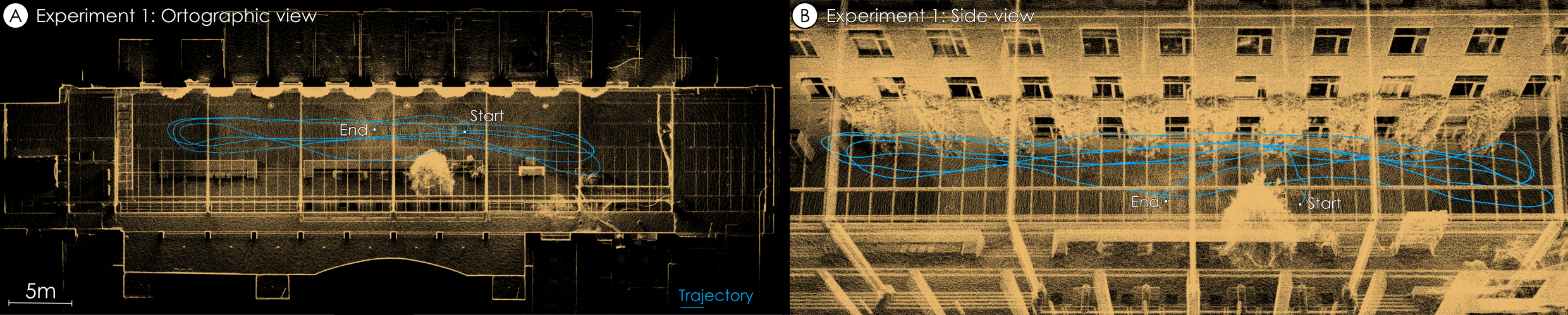}
\caption{Point cloud map and trajectories of experiment 1 from \cref{sec:evaluation:elektro}. \textbf{A}: Orthographic view; \textbf{B}: Side view.}
\label{fig:elektro}
\vspace{-1em}
\end{figure*}

\begin{table}[b]
    \setlength{\tabcolsep}{3.5pt}
    \caption{Quantitative results for experiment 1}
    \centering
\begin{tabular}{llcc|c}
 &  Metric&  Ours&  Baseline& LiO\\\cmidrule{2-5}
 \multirow{3}{*}{\rotatebox{90}{Timing}}&   Total [\unit{\second}]&  \textbf{21.88}&  46.50& 8.52\\
 &  Average [\unit{\milli\second}]&  \textbf{15.30}&  32.47& 5.97\\
 &  Per Iteration [\unit{\milli\second}]&  \textbf{4.13}&  4.84& 1.80\\
 \cmidrule{2-5}
 \multirow{2}{*}{\rotatebox{90}{Error}}&   ATE [\unit{\meter}]&  0.042 (0.020)&  0.041 (0.020)& 0.042 (0.020)\\
 &  RTE [\unit{\meter}]&  0.011 (0.010)&  0.011 (0.010)& 0.011 (0.010)\\ \cmidrule{2-5}
\end{tabular}\label{tab:ex1-elektro}
\\ 
\smallskip \raggedright \hspace{2.5em} Mean (standard deviation) ATE/RTE for translation. \\ 
\hspace{2.5em} Best \acl{rli} wall timings in bold.
\vspace{-1em}
\end{table}

\begin{figure}[t]
  \centering
  \includegraphics[width=0.9\linewidth]{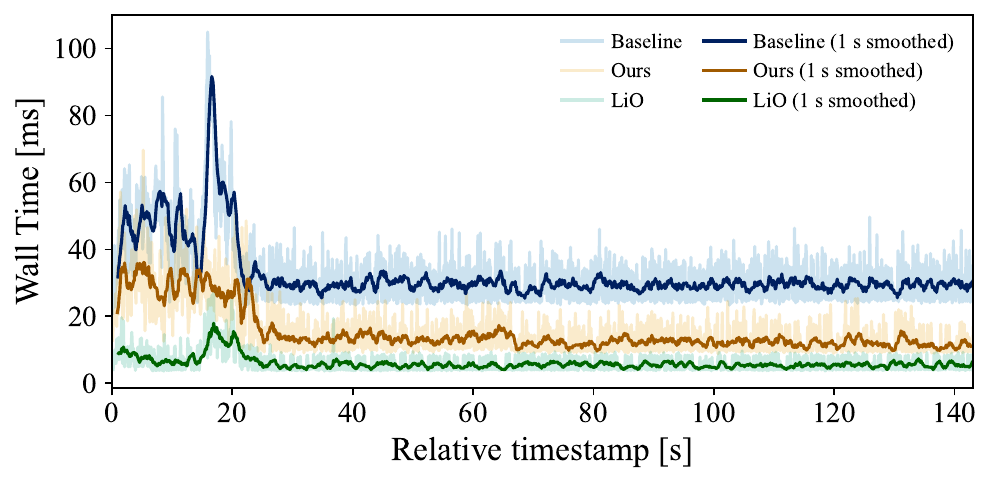}
  \caption{The factor graph optimization wall time per \qty{100}{\milli\second} for the baseline, proposed, and standard LiO methods on experiment 1 from \cref{sec:evaluation:elektro}. Overall, our method uses \qty{53}{\percent} less time than the baseline.}
  \label{fig:ex1-elektro-wall-time-plot}
  \vspace{-1em}
\end{figure}

In the Elektro Hall dataset, the platform is flown three rounds back and forth in a feature-rich indoor environment.

The environment is depicted in Fig. \ref{fig:fpf}~(E1) and the resulting point cloud map from the scan-to-map registration and integration is depicted in Fig. \ref{fig:fpf}~(P1) and Fig. \ref{fig:elektro}~(A-B). The maps built from our proposed method and the baseline are imperceptibly different, hence we only show the former. The quantitative results of experiment 1 are summarized in \cref{tab:ex1-elektro}. Note that the absolute and relative translational errors are, for all practical purposes, the same for both ours and the baseline methods. Simultaneously, the time used on factor graph optimization is \qty{53}{\percent} lower with our proposed method compared to the baseline. Individual factor graph iterations are also computationally cheaper, resulting in an average of \qty{15}{\percent} lower per-iteration time consumption. This is also visible in \cref{fig:ex1-elektro-wall-time-plot}, where the computational savings are evident in the deviation between the baseline and proposed approaches, with the baseline approach actually violating the \qty{100}{\milli\second} threshold for being real-time. It remains that the standard LiO approach is cheaper, however not meaningfully so as both approaches remain well under the limit to be real-time. Furthermore, this reduced optimization time comes at the cost of decreased robustness in degraded environments, as will be seen in \cref{sec:evaluation:fyllingsdalen}.

In summary, the experiment demonstrates a substantial reduction of computational load from using the proposed approach while leaving the localization accuracy unaffected. 

\subsection{Experiment 2: Fyllingdalen Bicycle Tunnel}\label{sec:evaluation:fyllingsdalen}

\begin{figure*}[t]
\centering
\vspace{0.5mm}
\begin{minipage}{\textwidth}
\centering
\includegraphics[width=\textwidth]{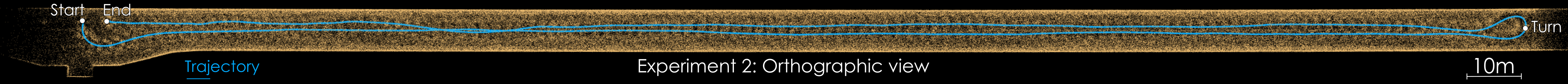}
\caption{Orthographic view of point cloud map and trajectory of experiment 2 from \cref{sec:evaluation:fyllingsdalen}.}
\label{fig:fyllingsdalen}
\end{minipage}
\par\addvspace{\floatsep}
\vspace{-1mm}
\begin{minipage}{\textwidth}
\centering
\includegraphics[width=\textwidth]{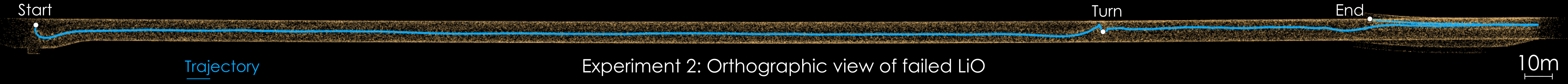}
\caption{Orthographic view of point cloud map and trajectory experiment 2 of \cref{sec:evaluation:fyllingsdalen} with LiO. LiO lacks resilience to the geometric degeneracy of the tunnel environment, resulting in gross errors in the estimated map and trajectory.}
\label{fig:fyllingsdalen-lio}
\end{minipage}
\vspace{-5mm}
\end{figure*}

\begin{table}[b]
\vspace{-1em}
    \setlength{\tabcolsep}{3.5pt}
    \caption{Quantitative results for experiment 2}
    \centering
\begin{tabular}{llcc|c}
 & Metric&  Ours&  Baseline& LiO\\\cmidrule{2-5}
 \multirow{3}{*}{\rotatebox{90}{Timing}} & Total [\unit{\second}]&  \textbf{40.49}&  91.75& 22.13\\
 & Average [\unit{\milli\second}]&  \textbf{18.92}&  42.83& 10.34\\
 & Per Iteration [\unit{\milli\second}]&  \textbf{3.83}&  4.58& 1.71\\
 \cmidrule{2-5}
    \multirow{2}{*}{\rotatebox{90}{Error}} & ATE [\unit{\meter}]&  0.494 (0.300)&  0.562 (0.281)& Failed \\
 & RTE [\unit{\meter}]&  0.025 (0.024)&  0.028 (0.024)& Failed\\ \cmidrule{2-5}
\end{tabular}
    \label{tab:ex2-fyllingsdalen}
\\
\smallskip \raggedright \hspace{4em} Mean (standard deviation) ATE/RTE for translation. \\ 
\hspace{4em} Best \acl{rli} wall timings in bold.
\vspace{-1em}
\end{table}

\begin{figure}[t]
  \centering
  \includegraphics[width=0.9\linewidth]{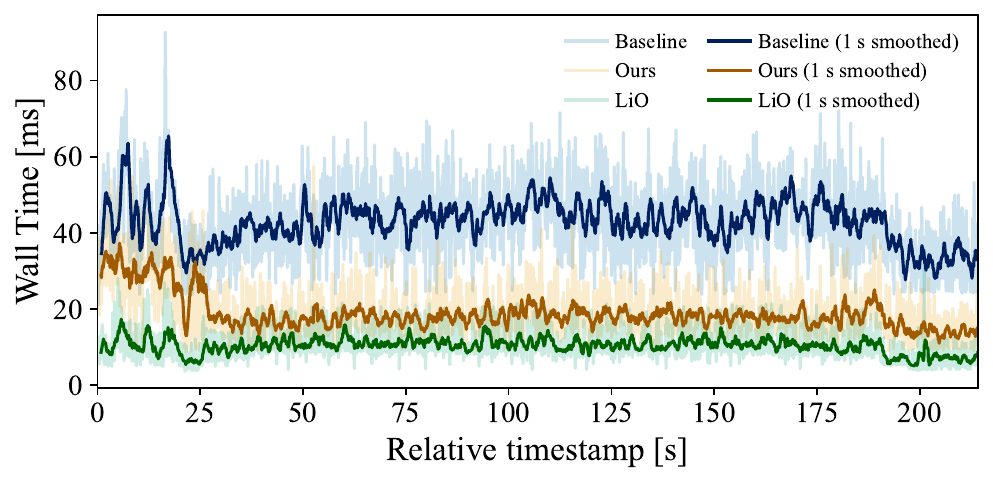}
  \caption{The factor graph optimization wall time per \qty{100}{\milli\second} for the baseline, proposed, and standard LiO methods on experiment 2 from \cref{sec:evaluation:fyllingsdalen}. Overall, our method uses \qty{56}{\percent} less time than the baseline.}
  \label{fig:ex2-fyllingsdalen-wall-time-plot}
\vspace{-1em}
\end{figure}

In the Fyllingsdalen tunnel dataset, the platform is flown approximately 250 meters along a geometrically uninformative environment before turning around and returning along the same path. The longitudinal translation along the direction of the tunnel is unobservable from the LiDAR alone due to a lack of salient geometric features. Despite the geometric degeneracy, the state estimates do not suffer as would be the case for LiDAR-only methods due to the Doppler measurements of the FMCW radar.

The environment is depicted in Fig. \ref{fig:fpf}~(E2) and the resulting point cloud map from the scan-to-map registration and integration is depicted in Fig. \ref{fig:fpf}~(P2) and Fig. \ref{fig:fyllingsdalen}. As in experiment 1, only the map from the proposed method is shown.

The quantitative results of experiment 2 are summarized in \cref{tab:ex2-fyllingsdalen}. The total time used for factor graph optimization is \qty{56}{\percent} lower with our proposed method compared to the baseline. Additionally, individual factor graph iterations are on average \qty{16}{\percent} cheaper. Simultaneously, the localization accuracy is similar for both approaches. As was the case for experiment 1 in \cref{sec:evaluation:elektro}, the computational savings here are substantial, as can be seen also in \cref{fig:ex2-fyllingsdalen-wall-time-plot}. LiO remains computationally cheaper, but its lack of resilience to the geometric degeneracy of the tunnel environment leads to gross localization errors. The localization failure is shown in Fig. \ref{fig:fyllingsdalen-lio}, where LiO significantly overshoots the longitudinal translation estimate from the starting point to the \qty{\sim250}{\meter} mark where the robot turns. LiO additionally fails to detect the turn, resulting in highly erroneous pose estimates.

For both ours and the baseline approaches, the translational errors are higher than in experiment 1. This is caused by the geometric degeneracy of the environment. While fusion of radar measurements improves the performance, it is still not at parity with purely LiDAR-inertial performance in geometrically well-informed environments. The absolute longitudinal translation is not measured directly by the LiDAR in all but the first part of the tunnel, and the estimated position therefore drifts.

Overall, experiment 2 supports the findings of experiment 1 in that the proposed approach substantially reduces the computational cost compared to the baseline without affecting localization accuracy.

\section{Conclusions}\label{sec:conclusions}
This paper presented a principled and computationally efficient method for asynchronous \acl{rli} factor graph-based smoothing. By introducing an IMU-preintegrated radar factor, we substantially reduce the number of states in the optimization problem by removing the need for dedicated radar state nodes in the graph. Experimental evaluation on a resource-constrained single-board computer demonstrated a reduction in total optimization time by up to \qty{56}{\percent} compared to a conventional baseline, without affecting localization accuracy. The results demonstrate that our proposed method can alleviate the computational burden of \acl{rli} sensor fusion, to the benefit of accurate sensor fusion on the computationally constrained embedded systems commonly found in mobile robotics.

\section*{ACKNOWLEDGMENT}
We would like to thank the Vestland Fylkeskommune for providing access to the Fyllingsdalen sykkeltunnel. 

\bibliographystyle{IEEEtran}
\bibliography{bibliography/IEEEabrv,bibliography/manual,bibliography/not-cleaned}

\end{document}

%% file: preamble.tex
\usepackage{graphics} 
\usepackage{graphicx}

\usepackage{epsfig} 
\usepackage{amsmath} 
\usepackage{amssymb}  

\usepackage{bm}
\usepackage{mathtools}

\usepackage{citesort}
\let\labelindent\relax
\usepackage{enumitem}

\usepackage{algorithm}
\usepackage[noend]{algpseudocode} 
\algnewcommand\AAND{\textbf{ and }}
\algnewcommand\Or{\textbf{ or }}

\usepackage{xcolor}
\usepackage{flushend}
\usepackage{url}
\usepackage[breaklinks]{hyperref}
\usepackage[capitalize]{cleveref}
\usepackage{booktabs}
\usepackage{makecell}
\usepackage{multirow}

\usepackage[nolist,nohyperlinks]{acronym}
\input{acronyms.tex}

\usepackage{tikz}
\usetikzlibrary{positioning, shapes.geometric, arrows.meta, decorations.pathreplacing, calligraphy}

\DeclareMathAlphabet{\pazocal}{OMS}{zplm}{m}{n}

\DeclareMathAlphabet{\mathpzc}{OT1}{pzc}{m}{it}

\usepackage{array}
\newcolumntype{C}[1]{>{\centering\arraybackslash}p{#1}}
\newcolumntype{M}[1]{>{\raggedright\arraybackslash}p{#1}}

\usepackage{array} 
\newcolumntype{L}[1]{>{\raggedright\let\newline\\\arraybackslash\hspace{0pt}}m{#1}}	
\newcolumntype{S}[1]{>{\centering\let\newline\\\arraybackslash\hspace{0pt}}m{#1}}
\newcolumntype{R}[1]{>{\raggedleft\let\newline\\\arraybackslash\hspace{0pt}}m{#1}}

\makeatletter
\renewcommand*{\@opargbegintheorem}[3]{\trivlist
  \item[\hskip \labelsep{\itshape #1\ #2}] \textit{(#3)}\ }
\makeatother

\usepackage[per-mode=symbol]{siunitx}
\usepackage{nth}

\def\hlfon{1} 
\def\todo#1{%
  \ifnum\hlfon=1
    \textcolor{red}{#1}%
  \else
    #1%
  \fi
}

%% file: acronyms.tex
\acrodef{gnss}[GNSS]{Global Navigation Satellite System}
\acrodef{icp}[ICP]{Iterative Closest Point}
\acrodef{uas}[UAS]{Unmanned Aerial Systems}
\acrodef{ransac}[RANSAC]{Random Sample Consensus}
\acrodef{slam}[SLAM]{Simultaneous Localization And Mapping}
\acrodef{fmcw}[FMCW]{Frequency Modulated Continuous Wave}
\acrodef{pca}[PCA]{Principal Component Analysis}
\acrodef{ekf}[EKF]{Extended Kalman Filter}
\acrodef{isam2}[iSAM2]{Incremental Smoothing and Mapping}
\acrodef{rio}[RIO]{Radar Inertial Odometry}
\acrodef{rmse}[RMSE]{Root Mean Square Error} 
\acrodef{ape}[APE]{Absolute Pose Error}
\acrodef{cfar}[CFAR]{Constant False Alarm Rate}
\acrodef{snr}[SNR]{Signal to Noise Ratio}
\acrodef{rcs}[RCS]{Radar Cross Section}
\acrodef{imu}[IMU]{Inertial Measurement Unit}
\acrodef{rli}[RLI]{Radar-LiDAR-Inertial}